\documentclass[11pt]{article}

\usepackage[final]{acl}

\usepackage{times}
\usepackage{latexsym}

\usepackage[T1]{fontenc}

\usepackage[utf8]{inputenc}

\usepackage{microtype}

\usepackage{inconsolata}

\usepackage{graphicx}
\usepackage{linguex} 
\usepackage{booktabs}
\usepackage{threeparttable}
\usepackage{multirow}

\usepackage{tcolorbox}
\tcbuselibrary{breakable}
\tcbset{
  promptbox/.style={
    colback=gray!15,
    colframe=gray!50,
    boxrule=0.3pt,
    arc=3pt,
    left=4pt, right=4pt, top=4pt, bottom=4pt,
    boxsep=3pt,
    breakable   
  }
}

\title{LLMBridge: An LLM Pipeline for End-to-end \\Referential Bridging Resolution in English}

\author{\textbf{Lauren Levine} \and 
\textbf{Amir Zeldes}  \\
  Georgetown University \\
  Department of Linguistics \\
  \texttt{\{lel76, amir.zeldes\}@georgetown.edu}}

\begin{document}
\maketitle
\begin{abstract}

In this paper, we introduce LLMBridge, a new LLM based system for the task of end-to-end referential bridging resolution in English. Our bridging resolution pipeline combines heuristic pre/post-processing with the natural language inference ability that comes from LLMs. We evaluate our bridging resolution pipeline on three datasets which have been used for referential bridging resolution evaluation in English: ISNotes, BASHI, and GUMBridge. Comparison to previous bridging resolution systems shows that the performance of LLMBridge surpasses previous state-of-the-art (SoTA) systems for all three datasets in the challenging End-to-end Evaluation Setting, as well as the Basic Bridging Resolution Evaluation Setting (gold bridging anaphor given). We also conduct a thorough error analysis of the LLMBridge performance, examining what varieties of bridging remain difficult for LLM based systems to identify. With this paper, we release the code for the LLMBridge pipeline.
\end{abstract}

\section{Introduction}


Bridging is an anaphoric phenomenon where the referent or type of a newly introduced entity is inferable due to its relationship with a previously introduced entity. Consider the following sentences:

\ex. There is \underline{a house}. \textbf{The door} is red.\footnote{Bridging anaphora are marked in bold face, and their associative antecedents are underlined.} \label{ex:opening}

In example \ref{ex:opening} above, we understand the newly introduced entity ``the door'' to specifically be the door of the aforementioned house, due to both the sequencing of the sentences and the semantic part-whole relationship that exists between the entities of ``house'' and ``door.'' In this bridging pair, ``the door'' is referred to as the bridging anaphor, and ``a house'' is its associative antecedent. Beyond such part-whole relations, the associative relationships that give rise to bridging manifest in a variety of different ways, including relative adjectives (\underline{a dog} $\rightarrow$ \textbf{a larger/different/other dog}), and prototypical associations (\underline{a library} $\rightarrow$ \textbf{the books}). 
Interpreting such implicit entity relations is necessary for various downstream NLP tasks, such as question-answering, controllable natural language generation and model output factuality verification. At present, it is unclear to what extent LLM systems track such implicit entity relations.

Bridging resolution is the task of automatically detecting bridging anaphora in natural language and resolving them back to their respective associative antecedents. While the task of bridging resolution has not received as much attention as other anaphoric phenomena like coreference resolution, it has received increased attention in recent years by being included in shared task datasets \cite{khosla-etal-2021-codi, yu-etal-2022-codi}, as well as independent efforts towards developing bridging datasets and bridging resolution systems \cite{kobayashi-ng-2020-bridging}. Previous bridging resolution systems have included rule based \cite{hou-etal-2014-rule, roesiger-etal-2018-bridging}, neural \cite{yu-poesio-2020-multitask, kobayashi-etal-2022-end}, and hybrid approaches \cite{kobayashi-etal-2022-constrained}. However, despite these attempts, bridging resolution has remained an extremely challenging NLP task. In the \textit{End-to-end Evaluation Setting}, SoTA systems do not exceed an F1 score of 40\% for anaphor recognition and do not exceed an F1 score of 30\% for anaphor resolution \cite{kobayashi-etal-2022-end, kobayashi-etal-2023-pairspanbert}. 

Previous work has shown the identification of bridging instances to be a very difficult task, even for human annotators, due to its highly subjective nature \cite{levine-zeldes-2025-subjectivity}. 
Thus far, there has been little exploration of LLMs' ability to reliably recognize bridging as a phenomenon. While recent work has included efforts to provide a benchmark/baseline in limited evaluation settings \cite{bu2025discotrack,levine-etal-2026-gumbridge}, there has been no previous attempt to leverage LLMs to perform the task of bridging resolution in the \textit{End-to-end Setting}. However, previous work framing bridging anaphora resolution as a QA task \cite{hou-2020-bridging}, as well as the recent bridging resolution baseline in \citet{levine-etal-2026-gumbridge}, suggest that LLM based query systems have the potential to improve upon previous bridging resolution systems. 

In this paper, we present LLMBridge, the first LLM based end-to-end pipeline for bridging resolution. 
We evaluate our bridging resolution pipeline on three English datasets for referential bridging: ISNotes \cite{markert-etal-2012-collective}, BASHI \cite{rosiger-2018-bashi}, and GUMBridge \cite{levine-etal-2026-gumbridge}, and we find that our pipeline surpasses previous SoTA systems in both the \textit{End-to-end} and \textit{Basic} bridging resolution evaluation settings. 
We provide code for reproducing the LLMBridge pipeline and evaluation, as well as code for preprocessing the evaluation datasets.\footnote{https://github.com/lauren-lizzy-levine/LLMBridge} We additionally provide detailed error analysis on the performance of LLMBridge, investigating the varieties of bridging that are easy/difficult for LLM based systems to identify.


\section{Background on Referential Bridging}

When bridging was first introduced to the literature, the ``bridge'' referred to the implicature that is constructed from the entity that is currently being processed back to a previous entity in a discourse that allowed for its interpretation \cite{clark-1975-bridging}. Initially, this covered a number of different entity relations, including identity relations, and bridging has since received a variety of theoretical treatments over time \cite{hawkins1978definiteness, prince1981toward, asher1998bridging, baumann2012referential, elazar-etal-2022-text}. This has led to a range of resources with different definitions of bridging and different annotation schemes.

\citet{roesiger-etal-2018-bridging} introduces the distinction between referential and lexical bridging as a means of describing the differences in the bridging definitions used by modern English bridging corpora. 
\textit{Referential bridging} refers to truly anaphoric instances of bridging where the referent or type of the bridging anaphor requires an antecedent to be interpretable, as in \ref{ex:referential}: 

\ex. She likes \underline{the house} because \textbf{the windows} are large.
\label{ex:referential}

On the other hand, \textit{lexical bridging} refers to lexical semantic relations between pairs of entities, such as part-whole or set-member relations, which may or may not be anaphoric, as in \ref{ex:lexical} where the antecedent is not strictly necessary for the interpretation:

\ex. I went to \underline{the United States} last month. My first stop was \textbf{Washington, DC}.
\label{ex:lexical}

Note that the referential meaning of ``Washington, DC'' in the example is recoverable without needing to refer back to ``the United States,'' though there is a semantic part-whole relation between the two. 

ISNotes, BASHI, and GUMBridge all take an information status based definition of bridging, and as a consequence, are all exclusively composed of instances of referential bridging.
The information status of an entity refers to the extent to which the entity is accessible to the reader/hearer of a discourse \citep{nissim-etal-2004-annotation}. An entity is either ``New'' information, where the entity is not yet known to the participant, or ``Given'' information, where the entity is recognized by the participant. However, when an entity is newly introduced but the referent is still inferable, it is said to be ``Accessible.'' In the case of a bridging anaphor, the entity is Accessible specifically due to its relation to a previous entity in the discourse, making it an anaphoric phenomenon.
 
In this paper, we focus on the task of bridging resolution for referential bridging. 
The ARRAU corpus \cite{poesio-artstein-2008-anaphoric, Uryupina2019AnnotatingAB} annotates related mentions that establish entity coherence through non-identity relations as bridging, rather than using an information status based definition. As such, ARRAU contains a mix of referential and lexical bridging, and we do not include ARRAU in our evaluation data. Table \ref{tab:corpora_stats} shows corpus statistics for the referential bridging corpora used in this study: ISNotes, BASHI, and GUMBridge. 

\begin{table}[t]
\centering
\resizebox{\columnwidth}{!}{%
\setlength{\tabcolsep}{4pt}
\begin{tabular}{l c c c c}
\toprule
  & \textbf{Genre(s)} 
 &  \textbf{Tokens} 
 & \shortstack{\textbf{Bridge}\\\textbf{Instances}}
 & \shortstack{\textbf{Bridge / 1k}\\\textbf{Tokens}} \\
\midrule
\textbf{ISNotes}      & WSJ news & 40k  & 916*  & 22.7 \\ [0.5ex]
\textbf{BASHI}        & WSJ news & 58k  & 459  & 7.9 \\ [0.5ex]
\textbf{GUMBridge}    & 24 genres & 291k & 5.7k & 19.6 \\
\bottomrule
\end{tabular}%
}

\begin{tablenotes}[flushleft]
\footnotesize
\item[] *Count includes instances of \texttt{mediated/bridging} and \texttt{mediated/comparative}.
\end{tablenotes}

\caption{Comparison of English referential bridging corpora used for evaluating bridging resolution systems.}
\label{tab:corpora_stats}
\end{table}

\section{Bridging Resolution}

\subsection{Task Definition}

Bridging resolution is the task of automatically recognizing the bridging anaphora in a text and resolving them back to their respective associative antecedents. 
The task of bridging resolution can be broken up and described in the following 3 subtasks:

\paragraph{Anaphor Recognition} Given the text of a discourse, identify the bridging anaphora present.

\paragraph{Anaphor Resolution} Given a bridging anaphor in a discourse, identify the associative antecedent which makes the referent or type of the anaphor inferable. When this subtask is done in isolation, it can also be referred to as \textit{antecedent selection}.

\paragraph{Subtype Classification} Given a bridging anaphor and antecedent pair in a discourse, select the sub-variety of bridging (e.g.~\textit{part-whole}) from a set of predefined semantic relations (a specific subtype schema must be assumed). 

\subsection{Evaluation Settings}

The evaluation of bridging resolution is commonly carried out in the three settings, each of which allows for a different amount of gold mention information to be present in the input data. A ``mention'' refers to a single occurrence of an entity (each of ``John'' and ``he'' in ``John said he’d be late''), and the gold mention information of a text refers to the token spans and conference cluster information of all coreferent and singleton mentions (entities mentioned only once) in a text. In the \textit{Basic Bridging Resolution Setting},\footnote{This setting has previously been referred to just as ``Bridging Resolution.'' We add the designation ``Basic'' to differentiate it from the general task name.} the system is given gold mention information and gold bridging anaphora, and the task is to resolve each bridging anaphor back to its respective associative antecedent (also called anaphor resolution/antecedent selection). In the \textit{Full Bridging Resolution Setting}, the system is given gold mention information, and the task is to both identify bridging anaphora and resolve them back to their respective associative antecedents in a discourse. In the \textit{End-to-end Bridging Resolution Setting}, the task is the same as in the \textit{Full Setting}, but the system is only given raw text as input.

Previous work on bridging resolution has primarily focused on the easier evaluation settings of \textit{Basic} and \textit{Full} bridging resolution, and less attention has been given to the more challenging \textit{End-to-end Setting}, despite it being the most realistic evaluation setting. 
As such, in this paper we focus on providing scores in the more realistic and challenging \textit{End-to-end Setting} for the English referential bridging resources: ISNotes, BASHI, and GUMBridge. In this evaluation setting, we report precision, recall, and F1-Score (P/R/F1) for Anaphor Recognition and Anaphor Resolution (the joint recognition of anaphor-antecedent pair). 
We also report scores in an altered version of the \textit{Basic Setting}, providing the gold anaphor as input, but not providing additional gold mention information. In this evaluation setting, we report Accuracy (antecedents correctly identified / total count of gold bridging anaphora). 
In both evaluation settings, we additionally report subtype classification scores for LLMBridge on the GUMBridge corpus. As GUMBridge allows for multiple subtype annotations, we report both Accuracy for exact match per bridging instance and P/R/F1 on predicting individual subtype annotations. 

We evaluate on the designated test split of GUMBridge, and the full datasets for ISNotes and BASHI (minus 5 documents from each corpus set aside for prompt development; see Appendix \ref{sec:appendix_prompts}), as they have no test splits. 
While we do not provide gold mention information in either of our evaluation settings, we use predicted mention and coreference information obtained from the available Stanza coref models trained on the GUM corpus\footnote{gum-nospeakers\_roberta-large-lora} \cite{Zeldes2017} (for GUMBridge) and the OntoNotes corpus\footnote{ontonotes-singletons\_roberta-large-lora} \cite{weischedel2011ontonotes} (for ISNotes and BASHI), which predict both coreferent and singleton mentions. 
Evaluation for bridging resolution can be done at the mention level (exact span match) or at the entity level (match to any member of the coreference cluster). As the English referential bridging resources all include data that has coreference annotations, we evaluate at the entity level, allowing any predicted antecedent that matches a (non-pronominal) mention of the gold antecedent entity cluster to be counted as a correct antecedent prediction.
We note that the annotations for the GUMBridge dataset were released online after the release of all models evaluated in this study, and, for ISNotes and BASHI, the document text and the bridging annotations are not provided in the same files of their respective data releases. As such, we believe the potential effect of data leakage to be limited.

\section{Previous Bridging Resolution Systems}

Since the rise in prominence of neural approaches following the introduction of Transformer models \cite{devlin-etal-2019-bert, vaswani2017attention}, there have been a number of attempts to address the task of bridging resolution with neural systems. 



\citet{kobayashi-etal-2022-end} give an overview of the performance of recent bridging resolution systems evaluated on ISnotes and BASHI in the more challenging \textit{End-to-end Setting}, in addition to the more commonly reported \textit{Full Setting}. Models adapted and evaluated in these experiments include \citet{roesiger-etal-2018-bridging}, \citet{yu-poesio-2020-multitask}, and \citet{kobayashi-ng-2021-bridging}. 
Most recently, \citet{kobayashi-etal-2023-pairspanbert} present \textsc{PairSpanBERT}, a pretrained model for bridging resolution based on \textsc{SpanBERT} \cite{joshi-etal-2020-spanbert}. Extending \citet{kobayashi-etal-2022-end} and other previous systems by leveraging \textsc{PairSpanBERT}, they evaluate on BASHI and ISNotes for both the \textit{End-to-end Setting} and \textit{Full Setting}. The system achieves near SoTA performance for the \textit{Full Setting}, with SoTA performance for the \textit{End-to-end Setting}. 

Taking a different approach, \citet{hou-2020-bridging} presents a system that re-frames the task of bridging resolution as a question answering task based on context. \citeauthor{hou-2020-bridging}'s BARQA (QA for Bridging Anaphora Resolution) system is designed to take in a segment of text and a question about an existing bridging anaphor in that text segment as inputs, and return the corresponding associative antecedent. Using this methodology, \citeauthor{hou-2020-bridging} reports SoTA performance on the ISNotes and BASHI corpora for the \textit{Basic Bridging Resolution Setting}. 


To serve as a comparison to the performance of LLMBridge, reported scores for the SoTA systems discussed above are shown in Table \ref{tab:end2end_results} for the \textit{End-to-end Setting}, and Table \ref{tab:reg_results} for the \textit{Basic Setting}. To additionally contextualize the performance of LLMBridge, we include here the inter-annotator agreement scores reported in the initial publications of the evaluation resources: \citet{markert-etal-2012-collective} reports Cohen’s $\kappa$ of three annotators on bridging anaphor recognition for ISNotes: annotators A-B: 70.8, annotators A-C: 60.6, annotators B-C: 62.3. \citet{rosiger-2018-bashi} reports anaphor recognition agreement of 70.9\%, and 59.3\% agreement for the overall links on BASHI. \citet{levine-etal-2026-gumbridge} reports an anaphor recognition mutual F1 of 62\% and an antecedent selection accuracy of 84\% on GUMBridge.






\section{LLMBridge Pipeline}

LLMBridge is an LLM based pipeline for the task of bridging resolution. It combines heuristic pre/post processing with backend LLM queries to create a robust bridging resolution system for referential bridging. The pipeline handles three subtasks of bridging resolution: (1) anaphor recognition, (2) anaphor resolution, and (3) subtype classification. Our system provides code to run the subtasks individually (as in the \textit{Basic Bridging Resolution Setting}), where gold input is given for the antecedent selection and subtype classification tasks, or as a full end-to-end pipeline, where the output of one task is used as the input for the next (as in the \textit{End-to-end Setting}).


\subsection{Subtask Operationalization}

The following describes how we operationalize the subtasks of bridging resolution to be accomplished via LLM query. In our prompt design, we follow an information status based, referential definition of bridging, where a bridging anaphor must be a newly introduced entity, and it must be inferable due to an anaphoric, associative (non-identity) relation to a previous entity in the discourse. Complete prompt templates are included in Appendix \ref{sec:appendix_prompts}. The buffer/context window sizes described in the subtask operationalizations below are all a configurable part of the preprocessing. 


\paragraph{Anaphor Recognition}

For each sentence in the input text, the backend LLM is queried and asked to identify any bridging anaphora in the sentence. The query provides a definition of bridging anaphora, instructions for the anaphor recognition task, examples, the text of the sentence being asked about with up to 5 tokens of context buffer to the left and right of the sentence, and a back context of up to 150 tokens.\footnote{An analysis of the train/dev partitions has shown > 90\% of associative antecedents fall within that window.} The LLM is instructed to return a list of bridging anaphora. 

\paragraph{Anaphor Resolution}

For each bridging anaphor predicted by the system, the backend LLM is queried and asked to identify the associative antecedent. The query provides a definition of bridging anaphora, instructions for the antecedent selection task, examples, and the text of the sentence containing the predicted anaphor (marked in double curly brackets: \{\{ \}\}) with up to 150 tokens of context to the left of the sentence. The LLM is expected to return the identified antecedent entity, or if no antecedent is identified, it is instructed to return the string ``no antecedent.'' In the \textit{Basic Bridging Resolution Setting}, the anaphor resolution subtask is run with the sentence containing the gold anaphor and 150 tokens of back context as input.

\paragraph{Subtype Classification (GUMBridge only)}
For each bridging anaphor-antecedent pair predicted by the system (candidate anaphora where ``no antecedent'' is predicted are filtered out), the backend LLM is asked to classify the bridging subtype of the pair. The query provides a definition of bridging anaphora, instructions for the subtype classification task (including details on the 11 bridging subtypes in the GUMBridge annotation schema\footnote{Details of the subtype varieties in GUMBridge are included in Appendix \ref{sec:appendix_subtype}}), examples, and the sentence containing the anaphor with 150 tokens of back context (with the anaphor marked in double curly brackets: \{\{ \}\} and the antecedent marked in asterisks: * *). In the \textit{Basic Bridging Resolution Setting}, the LLM receives the individual sentences containing the anaphor and antecedent with 10 tokens of buffer context to the left and right. The LLM is instructed to return all of the bridging subtypes applicable to the pair.


\subsection{LLM Model Selection}


By design, any LLM can serve as the backend to be queried in the LLMBridge pipeline. We provide scores for one closed SoTA model, \texttt{Gemini-3.1-pro-preview}, and two smaller open models, \texttt{meta-llama/llama-3.3-70b-instruct} and \texttt{Qwen2.5-7B-Instruct}, in order to demonstrate the competence of different sized language models on bridging resolution. 
We also provide scores for \texttt{Qwen2.5-7B-Instruct-FT}, a version of \texttt{Qwen2.5-7B-Instruct} which we fine-tune using low-rank adapters (LoRA; \citealp{hu2022lora}) on the GUMBridge train data (213k tokens, 4k bridging instances), which is the largest English corpus for referential bridging and includes annotations for bridging subtypes. We fine-tune the model for 3 epochs using a learning rate of 3e-4, with low-rank adapters of rank 8 and a LoRA alpha of 16. We release the \texttt{Qwen2.5-7B-Instruct-FT} bridging fine-tuned model with the LLMBridge pipeline code. We include this model in our experiments to investigate the gains that fine-tuning can bring for smaller LLMs, even with limited data and compute.

\subsection{Pre/Post Processing}

To boost the performance of our pipeline, we include a number of heuristics which are based on additional linguistic annotations, including coreference/mention, part-of-speech, and dependency syntax (UD; \citealp{10.1162/coli_a_00402}). If the additional linguistic annotations are not available, predicted annotations can be used (we provide code to enable this within our pipeline, leveraging Stanza, \citealt{qi-etal-2020-stanza}), or the pipeline can run without the pre/post processing. We report scores using predicted annotations from Stanza during the pre/post processing for the \textit{End-to-end} and \textit{Basic} evaluation settings. To examine the impact of the following heuristics on the performance of LLMBridge, in Appendix \ref{sec:processing_comparison} we provide scores for running the \texttt{Qwen2.5-7B-Instruct-FT} model without pre/post processing in both evaluation settings. 

\paragraph{Adjusting Responses to Attested Mention Spans}

We adjust the span of the LLM's predicted response to match an actual, attested (or Stanza predicted) mention span for the data in order to improve precision. We adjust the predicted anaphor returned by the LLM to be the shortest existing mention span which contains the predicted anaphor. If there is no such existing mention, we check whether there are any existing mention spans which are contained within the predicted anaphor, and, if so, adjust the anaphor span to be the longest such mention. If there is still no such mention, the predicted anaphor is rejected. We also do this adjustment of the predicted span for the antecedent selection.

\paragraph{Suggesting Candidate Bridging Anaphora}

In order to improve recall, we also use heuristics to identify possible candidates for bridging anaphora and make additional queries to the LLM backend to get specific judgments on them. The entities we highlight include entities containing comparative adjectives (e.g., ``a smaller dog''), entities containing any keyword from a list of relative or temporal markers (e.g., ``another'', ``others'', ``different'', ``following'', ``yesterday'', etc.), and any two-token entity where the first token is a definite determiner (e.g, ``the door'').
The additional LLM query for anaphor recognition with selected candidates is done immediately following the regular call for anaphor recognition, and then the results are combined (unioned) into a single list of predicted bridging anaphora that are passed onto the next stage of the pipeline. The prompt is largely the same as the regular anaphor recognition prompt, but it includes the instruction to choose from a provided candidate list (see Appendix \ref{sec:appendix_prompts}).

\paragraph{Filtering Coreference from Predicted Bridging Anaphora}

In the anaphoric, information status based definition of referential bridging, a bridging anaphor must be a newly introduced entity in the discourse. As such, any entity with a previous mention is structurally prohibited from being a bridging anaphor. Therefore, in order to improve precision on referential bridging, we filter out any predicted bridging anaphora that are subsequent mentions in a coreference chain. 


\section{Results}

\begin{table*}[h]
\centering
\resizebox{0.8\linewidth}{!}{%
\begin{tabular}{ccccccc}
\toprule
\multicolumn{7}{c}{\textbf{\Large{End-to-end Bridging Resolution Setting}}} \\ \midrule
\multirow{2}{*}{Model} & \multicolumn{3}{c}{Recognition} & \multicolumn{3}{c}{Resolution} \\
 & P & R & \multicolumn{1}{c|}{F1} & P & R & F1 \\ \midrule
\multicolumn{7}{c}{\textbf{GUMBridge}} \\ \midrule
\multicolumn{1}{l}{Gemini} & 43.3 $\pm$ 1.6 & 60.6 $\pm$ 4.4 & \textbf{50.4} $\pm$ 0.2 & 26.4 $\pm$ 2.0 & 36.9 $\pm$ 1.3 & \textbf{30.7} $\pm$ 0.9 \\
\multicolumn{1}{l}{Llama} & 22.4 $\pm$ 0.7 & 56.3 $\pm$ 1.9 & 32.0 $\pm$ 0.4 & 8.5 $\pm$ 0.6 & 21.4 $\pm$ 0.2 & 12.2 $\pm$ 0.7 \\
\multicolumn{1}{l}{Qwen-FT} & 33.7 $\pm$ 1.1 & 38.7 $\pm$ 1.1 & 36.0 $\pm$ 1.1 & 13.3 $\pm$ 0.8 & 15.2 $\pm$ 0.9 & 14.2 $\pm$ 0.8 \\
\multicolumn{1}{l}{Qwen} & 25.1 $\pm$ 0.4 & 36.2 $\pm$ 5.4 & 29.6 $\pm$ 1.6 & 2.1 $\pm$ 0.2 & 3.0 $\pm$ 0.3 & 2.5 $\pm$ 0.0\\ \midrule
\multicolumn{7}{c}{\textbf{ISNotes}} \\ \midrule
\multicolumn{1}{l}{Gemini} & 37.3 $\pm$ 1.7 & 57.2 $\pm$ 2.1 & \textbf{45.1} $\pm$ 0.6 & 23.7 $\pm$ 2.2 & 36.3 $\pm$ 0.2 & \textbf{28.7} $\pm$ 1.6 \\
\multicolumn{1}{l}{Llama} & 19.0 $\pm$ 0.0 & 48.6 $\pm$ 3.1 & 27.3 $\pm$ 0.5 & 7.1 $\pm$ 0.8 & 18.1 $\pm$ 0.8 & 10.2 $\pm$ 1.0 \\
\multicolumn{1}{l}{Qwen-FT} & 27.6 $\pm$ 0.2 & 31.3 $\pm$ 0.6 & 29.4 $\pm$ 0.1 & 9.4 $\pm$ 0.3 & 10.7 $\pm$ 0.6  & 10.0 $\pm$ 0.4 \\ 
\multicolumn{1}{l}{Qwen} & 18.5 $\pm$ 0.2 & 23.7 $\pm$ 4.2 & 20.7 $\pm$ 1.8 & 2.7 $\pm$ 0.5 & 3.5 $\pm$ 0.0 & 3.0 $\pm$ 0.3 \\ \midrule
\multicolumn{1}{l}{\citet{kobayashi-etal-2023-pairspanbert}} & 40.2 & 39.5 & 39.9 & 26.4 & 25.9 & 26.2 \\ \midrule
\multicolumn{7}{c}{\textbf{BASHI}} \\ \midrule
\multicolumn{1}{l}{Gemini} & 22.7 $\pm$ 0.4 & 73.6 $\pm$ 5.1 & \textbf{34.7} $\pm$ 0.2 & 13.2 $\pm$ 1.2 & 43.0 $\pm$ 0.3 & \textbf{20.2} $\pm$ 1.3 \\
\multicolumn{1}{l}{Llama} & 9.5 $\pm$ 0.0 & 63.8 $\pm$ 2.9 & 16.5 $\pm$ 0.1 & 3.7 $\pm$ 0.2 & 24.9 $\pm$ 0.2 & 6.4 $\pm$ 0.3 \\
\multicolumn{1}{l}{Qwen-FT} & 16.4 $\pm$ 0.9 & 47.7 $\pm$ 3.6 & 24.4 $\pm$ 1.5 & 4.5 $\pm$ 0.6 & 13.3 $\pm$ 1.5 & 6.8 $\pm$ 0.9 \\ 
\multicolumn{1}{l}{Qwen}  & 10.3 $\pm$ 0.3 & 38.8 $\pm$ 5.6 & 16.2 $\pm$ 0.2 & 0.9 $\pm$ 0.3 & 3.2 $\pm$ 0.7 & 1.4 $\pm$ 0.5 \\ \midrule
\multicolumn{1}{l}{\citet{kobayashi-etal-2023-pairspanbert}} & 43.0 & 25.6 & 32.1 & 25.4 & 14.3 & 18.3 \\
\bottomrule
\end{tabular}%
}


\caption{LLMBridge and previous system results on evaluation datasets in the \textit{End-to-end Setting}.}
\label{tab:end2end_results}
\end{table*}

We evaluate LLMBridge on the aforementioned English referential bridging datasets: ISNotes, BASHI, and GUMBridge. We provide scores for the two run average (with standard deviation) on the datasets for the \textit{End-to-end Bridging Resolution Setting} in Table \ref{tab:end2end_results}, and the altered \textit{Basic Bridging Resolution Setting} (gold bridging anaphor, but no gold mention information) in Table \ref{tab:reg_results}. When available, we also include previous SoTA bridging resolution scores reported on the datasets for comparison. In Table \ref{tab:subtype_results}, we provide bridging subtype classification scores on GUMBridge in both evaluation settings. The highest scores for the F1 and Accuracy metrics are shown in \textbf{bold}.

\begin{table}[]
\centering
\resizebox{\columnwidth}{!}{%
\begin{tabular}{cccc}
\toprule
\multicolumn{4}{c}{\textbf{\Large{Basic Bridging Resolution Setting}}} \\ \midrule
Model & \textbf{GUMBridge} & \textbf{ISNotes} & \textbf{BASHI} \\ \midrule
\multicolumn{1}{l}{Gemini} & \textbf{56.7} $\pm$ 0.1 & \textbf{58.4} $\pm$ 0.4 & \textbf{56.4} $\pm$ 0.5 \vspace{0.1cm}\\
\multicolumn{1}{l}{Llama} & 35.3 $\pm$ 0.4 & 34.7 $\pm$ 0.5 & 37.3 $\pm$ 0.7 \vspace{0.1cm} \\ 
\multicolumn{1}{l}{Qwen-FT} & 29.8 $\pm$ 0.1 & 28.9 $\pm$ 0.1 & 24.6 $\pm$ 0.5 \vspace{0.1cm} \\ 
\multicolumn{1}{l}{Qwen} & 11.8 $\pm$ 0.5 & 9.8 $\pm$ 0.1 & 10.2 $\pm$ 0.9 \\ \midrule
\multicolumn{1}{l}{\citet{levine-etal-2026-gumbridge}} & 49.6 & - & - \vspace{0.1cm} \\
\multicolumn{1}{l}{\citet{hou-2020-bridging}} & - & 50.1 & 38.7 \\
\bottomrule
\end{tabular}%
}


\caption{LLMBridge and previous system results on evaluation datasets for antecedent selection accuracy in the \textit{Basic Setting}.}
\label{tab:reg_results}
\end{table}

\begin{table}[]
\centering
\resizebox{\columnwidth}{!}{%
\begin{tabular}{ccccc}
\toprule
\multicolumn{5}{c}{\textbf{\Large{GUMBridge Subtype Classification}}} \\ \midrule
Model & P & R & F1 & Acc \\ \midrule
\multicolumn{5}{c}{\textbf{End-to-end Bridging Resolution Setting}} \\ \midrule
\multicolumn{1}{l}{Gemini} & 76.9 $\pm$ 1.5 & 46.9 $\pm$ 3.3 & \textbf{58.6} $\pm$ 3.0 & \textbf{44.8} $\pm$ 3.0 \vspace{0.1cm} \\
\multicolumn{1}{l}{Llama} & 42.4 $\pm$ 2.5 & 37.7 $\pm$ 3.6 & 39.9 $\pm$ 3.1 & 35.5 $\pm$ 2.9 \vspace{0.1cm} \\
\multicolumn{1}{l}{Qwen-FT} & 57.4 $\pm$ 2.1 & 22.1 $\pm$ 1.4 & 32.0 $\pm$ 1.8 & 19.2 $\pm$ 0.8 \vspace{0.1cm} \\
\multicolumn{1}{l}{Qwen} & 40.2 $\pm$ 1.2 & 13.3 $\pm$ 2.3 & 19.9 $\pm$ 2.8 & 11.4 $\pm$ 1.8 \\ \midrule
\multicolumn{5}{c}{\textbf{Basic Bridging Resolution Setting}} \\ \midrule
\multicolumn{1}{l}{Gemini} & 78.7 $\pm$ 0.2 & 79.0 $\pm$ 0.9 & \textbf{78.9} $\pm$ 0.5 & \textbf{77.1} $\pm$ 1.0 \vspace{0.1cm} \\
\multicolumn{1}{l}{Llama} & 44.8 $\pm$ 0.2 & 63.5 $\pm$ 1.8 & 52.6 $\pm$ 0.5 & 61.0 $\pm$ 2.0 \vspace{0.1cm} \\
\multicolumn{1}{l}{Qwen-FT} & 61.7 $\pm$ 0.9 & 60.7 $\pm$ 0.7 & 61.2 $\pm$ 0.8 & 57.3 $\pm$ 1.3 \vspace{0.1cm} \\
\multicolumn{1}{l}{Qwen} & 37.8 $\pm$ 0.3 & 34.7 $\pm$ 0.3 & 36.2 $\pm$ 0.3 & 31.9 $\pm$ 0.1 \\
\bottomrule
\end{tabular}%
}
\caption{LLMBridge subtype classification results on GUMBridge in \textit{End-to-end} and \textit{Basic} evaluation settings.}
\label{tab:subtype_results}
\end{table}

Overall, we see that leveraging \texttt{Gemini-3.1-pro-preview} as the LLM backend of the pipeline yields the highest scores across all datasets and evaluation settings for LLMBridge. This is expected as \texttt{Gemini-3.1-pro-preview} is by far the largest model tested here, though it is notable that the scores are still not high. We observe an overall increase in scores moving from the results of \texttt{Qwen2.5-7B-Instruct} to that of \texttt{Llama-3.3-70B-Instruct} to that of \texttt{Gemini-3.1-pro-preview}. This demonstrates the expected trend that larger, more powerful chain-of-thought (CoT) reasoning models give stronger performance on high level discourse tasks like bridging resolution.

When comparing the different evaluation settings, we expect the \textit{End-to-end Bridging Resolution Setting} to be more challenging than the \textit{Basic Bridging Resolution Setting}. We see this reflected in the performance of LLMBridge across the different models and datasets, with scores increasing from the \textit{End-to-end} to \textit{Basic} Setting. Taking the \texttt{Gemini-3.1-pro-preview} GUMBridge scores as an illustrative example, we can see that the accuracy score for antecedent selection in the \textit{Basic Setting} is 56.7, while the F1 scores for anaphor recognition and resolution are 50.4 and 30.7 in the \textit{End-to-end Setting}. The scores for other datasets and models follow the same pattern.

\begin{figure}
  \centering
  \includegraphics[width=1.0\linewidth]{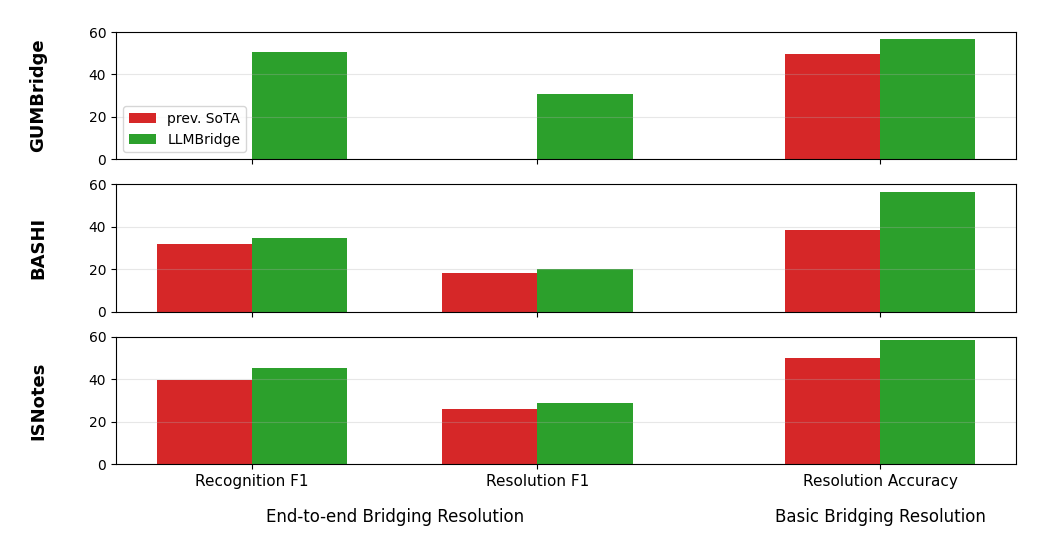}
  \caption{LLMBridge (Gemini Backend) vs. Previous SoTA Results on Referential Bridging Datasets}
  \label{fig:LLMBridge_SoTA}
\end{figure}

Comparing the best performance of LLMBridge (\texttt{Gemini-3.1-pro-preview} backend) to the reported scores from previous systems, we see that we achieve SoTA results on all datasets in both the \textit{End-to-end} and \textit{Basic} evaluation settings. In Figure \ref{fig:LLMBridge_SoTA}, we provide a comparison of the LLMBridge scores (\texttt{Gemini} backend) to previous SoTA results on the referential bridging datasets.
For BASHI, we achieve SoTA results in the \textit{Basic Setting} with an antecedent selection accuracy of 56.4 (+17.7 previous SoTA) and in the \textit{End-to-end Setting} with an anaphor recognition F1 of 34.7 (+2.6) and an anaphor resolution F1 of 20.2 (+1.9). For ISNotes, we achieve SoTA results in the \textit{Basic Setting} with an antecedent selection accuracy of 58.4 (+8.3), and in the \textit{End-to-end Setting} with an anaphor recognition F1 of 45.1 (+5.2) and an anaphor resolution F1 of 28.7 (+2.5). For GUMBridge we achieve SoTA on antecedent selection accuracy in the \textit{Basic Setting} at 56.7 (+7.1). We also provide the first \textit{End-to-end} bridging resolution scores on the GUMBridge dataset. The anaphor recognition and anaphor resolution F1-Scores we report for GUMBridge, 50.4 and 30.7 respectively, are the highest F1-Scores ever reported on any English bridging dataset in the \textit{End-to-end Setting}.

When we compare between the LLMBridge results for \texttt{Qwen2.5-7B-Instruct} and \texttt{Qwen2.5-7B-Instruct-FT}, we see that there is a sizable increase in performance across evaluation settings and datasets. 
In the \textit{End-to-end Setting} we see an avg. $\Delta$ F1 across datasets of +7.8 for anaphor recognition and +8.0 for anaphor resolution. In the \textit{Basic Setting} there is an avg. $\Delta$ accuracy of +17.2 for anaphor resolution. 
We also see that this jump in performance is large enough to be competitive across datasets with the performance of the larger \texttt{Llama-3.3-70B-Instruct} in the \textit{End-to-end Setting}.
With regard to subtype classification (GUMBridge only),  we also see an increase in performance, with a $\Delta$ F1 of +12.1 in the \textit{End-to-end Setting} and a $\Delta$ F1 of +25.0 in the \textit{Basic Setting}. 
This increase in performance across subtasks illustrates the utility of fine-tuning an LLM for bridging resolution task, even when the training data and compute resources are limited. If resource availability prevents simply scaling up model size, fine-tuning a smaller model can still result in meaningful gains.


Finally, looking at the GUMBridge subtype classification scores in Table \ref{tab:subtype_results}, we see that the performance is considerably higher in the easier \textit{Basic Setting} than in the harder \textit{End-to-end Setting} where classification is a downstream task. As GUMBridge allows for multiple subtype annotations on an instance of bridging, we report P/R/F1 on detection of individual subtype labels. In the \textit{Basic Setting}, we achieve an F1 score of 78.9. For exact match on subtype labels, we achieve an accuracy score of 77.1 in the \textit{Basic Setting}. These are the first subtype classification scores for GUMBridge reporting both P/R/F1 and accuracy, and, overall, such scores indicate moderate success for the task of bridging subtype classification. 







\section{Error Analysis}

In this section, we conduct error analysis on the predictions from the best performing configuration of the LLMBridge pipeline (a single run from the \texttt{Gemini-3.1-pro-preview} backend). We conduct this error analysis in order to determine what sub-varieties of referential bridging remain difficult for an LLM based bridging resolution system like LLMBridge. We explore the influence of anaphor-antecedent pair distance and bridging subtype on the ability of the LLMBridge pipeline to identify instances of bridging anaphora. We additionally examine LLMBridge's performance on subtype categorization, looking at which subtypes are easier/harder for the system to classify. As GUMBridge is the only dataset with subtype labels, the subtype related investigations exclusively analyze the predictions on GUMBridge test. Details on the GUMBridge subtype categorization schema are given in Appendix \ref{sec:appendix_subtype}.

\begin{figure}
  \centering
  \includegraphics[width=1.0\linewidth]{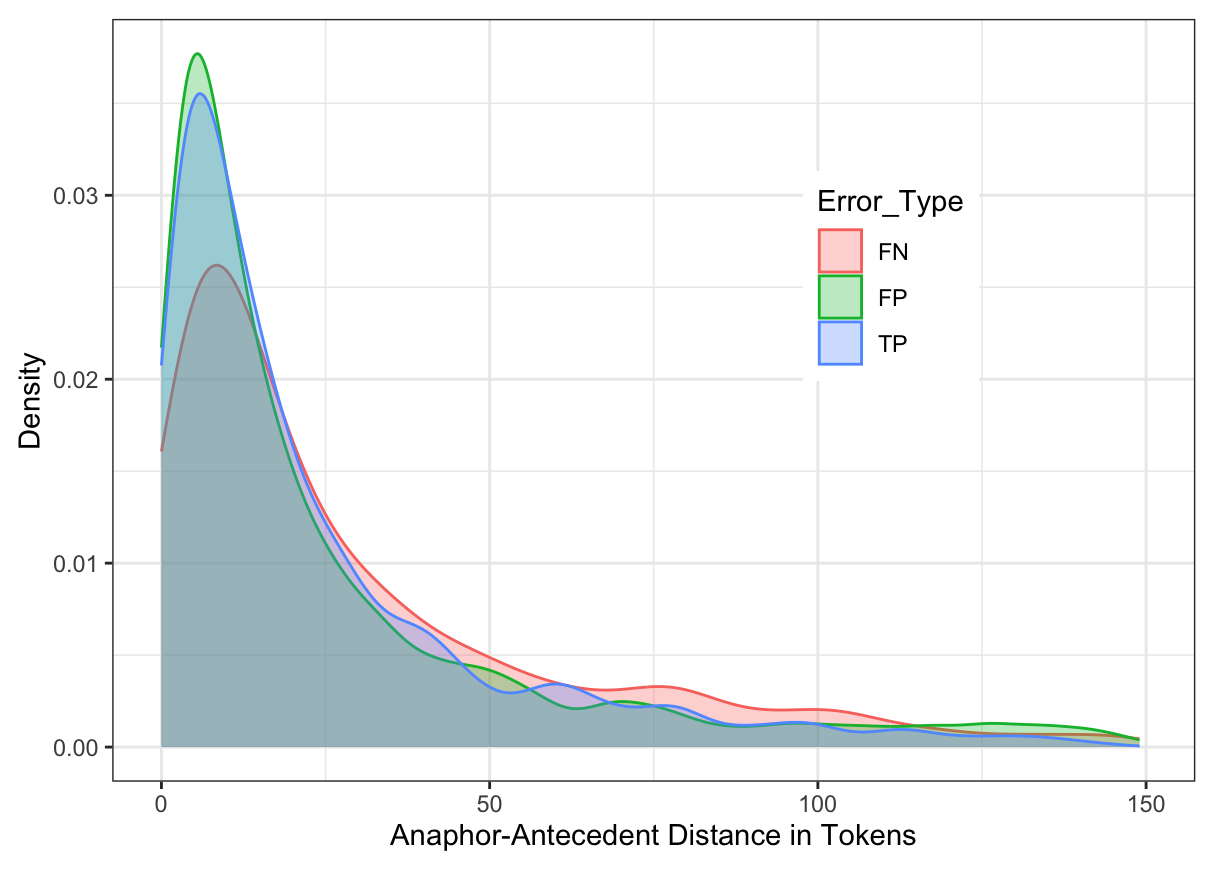}
  \caption{Distribution of anaphor-antecedent distances for bridging pairs with error types of False Negative (red), False Positive (green), and True Positive (blue) for LLMBridge (Gemini backend) performance on anaphor recognition.}
  \label{fig:dist_density}
\end{figure}

In Figure \ref{fig:dist_density}, we show the distribution of anaphor-antecedent distances for bridging pairs in ISNotes, BASHI, and GUMBridge grouped by whether LLMBridge correctly identified the bridging anaphor (True Positive; TP), failed to identify the bridging anaphor (False Negative; FN), or mistakenly identified a non-bridging entity as a bridging anaphor (False Positive; FP). Because the context window given to LLMBridge for anaphor detection and antecedent selection was only 150 tokens (> 90\% of bridging instances in the data are < 150 tokens apart), we first filter out long distance instances of bridging more than 150 tokens apart. Looking at the density curves in Figure \ref{fig:dist_density}, we see that a larger part of the density curve for False Negatives covers higher token distances when compared with the curves for True Positives and False Positives. Kolmogorov-Smirnov tests comparing the FN distribution with the TP and FP distributions confirm that LLMBridge is more likely to fail to identify instances of bridging when the anaphor and antecedent are further apart (Holm adjusted p-values both < 0.01). In other words, instances of long distance bridging are more difficult for the system to identify. It is notable that this trend is apparent even when constrained to examining instances of bridging that occur within 150 tokens. We also note the FP distribution is not significantly different from the TP distribution, indicating that although FP predictions are wrong, they are being made in a reasonable distance range.

\begin{figure}
  \centering
  \includegraphics[width=.9\linewidth]{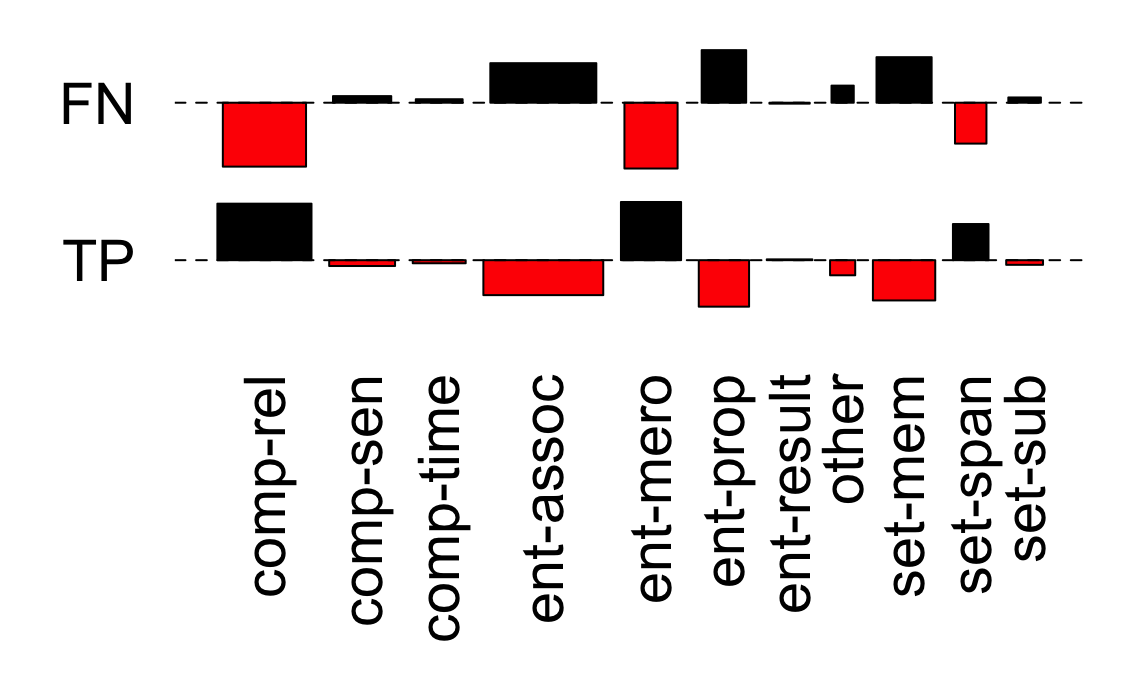}
  \caption{$\chi^2$ residuals: True Positive/False Negative LLMBridge predictions (\texttt{Gemini} backend) on anaphor recognition and gold bridging subtype label. ($\chi^2$ = 28.899, df = 10, p < 0.01)}
  \label{fig:ana_subtype}
\end{figure}

In Figure \ref{fig:ana_subtype}, we show an association plot of the residuals from a $\chi^2$ test for gold bridging instances that were correctly identified (TP) or missed (FN) in LLMBridge's performance on the task of anaphor recognition and the gold subtype labels of those bridging instances. Looking at the TP row of the association plot, we can see that LLMBridge is more able to identify bridging anaphora labeled with \textsc{comparison-relative} or \textsc{entity-meronomy}.
The \textsc{comparison-relative} instances being easier to identify likely reflects the tendency for this subtype to have overt markers (e.g., comparative markers such as ``other'' or ``another''). \textsc{entity-meronomy} may be easier to recognize due to the part-whole relations being more lexically inferable from the entities alone (e.g., \underline{a shady garden bed} → \textbf{the soil}). Looking at the FN row, we can see that LLMBridge is more likely to miss bridging instances labeled with \textsc{entity-associative}, \textsc{entity-property}, or \textsc{set-member}. It is unsurprising that the \textsc{entity-associative} subtype is difficult, as associative entity relations comprise the broadest sub-variety, covering a variety of implicit relations which lack overt markers, such as relational nouns (e.g., \underline{a business} → \textbf{the customer}), implicit arguments (e.g., \underline{a murder} → \textbf{the victim}), and prototypical associations (e.g., \underline{a wedding} → \textbf{the reception}). For the subtypes \textsc{entity-property} and \textsc{set-member}, the relations may also be more abstract (e.g., intangible qualities and class-instance relations) or more dependent on a wider context for understanding (i.e., it may be difficult to infer a set relation just from the anaphor itself). 



\begin{figure}
  \centering
  \includegraphics[width=1.0\linewidth]{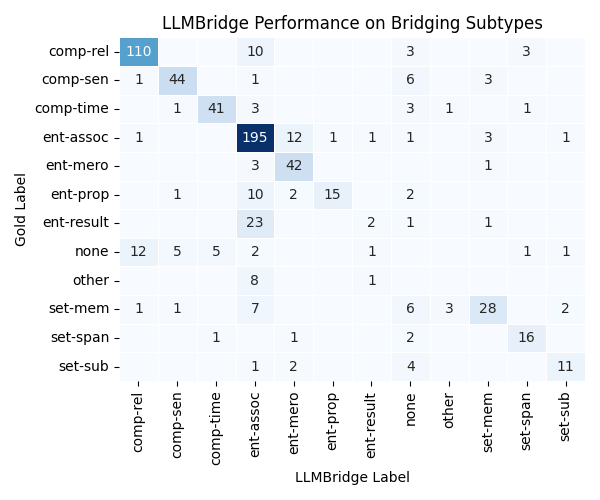}
  \vspace{-4pt}
  \caption{Confusion matrix of LLMBridge (\texttt{Gemini} backend) predicted subtype labels and gold labels.}
  \vspace{-6pt}
  \label{fig:subtype_cm}
\end{figure}


In Figure \ref{fig:subtype_cm}, we provide a confusion matrix of the bridging subtype labels assigned by LLMBridge (\texttt{Gemini-3.1-pro-preview} backend) and the gold labels from the test set of GUMBridge. In order to have more data to analyze, we consider the predictions from \textit{Basic} evaluation setting, where the gold bridging pair is provided. Looking at Figure \ref{fig:subtype_cm}, we see that while there is overlap for a variety of different subtypes, most confusions occur only 5 or less times. If we look at the \textit{none} row of the gold labels axis, we can see the counts of instances where LLMBridge extraneously generated an additional label. We can see that these hallucinated labels are primarily \textsc{comparison} relations. And while it is somewhat common for \textsc{comparison} labels to co-occur with other subtype labels (e.g., \underline{a book} → \textbf{another one} being an instance of both \textsc{comparison-relative} and \textsc{comparison-sense}), it appears that LLMBridge over-generates such cases when only a single label is actually applicable. Looking at Figure \ref{fig:subtype_cm}, we can also see that the subtypes with the greatest overlap are \textsc{entity-resultative} and \textsc{entity-associative}, with 23 instances of \textsc{resultative} being mistaken by the system for instances of \textsc{associative}. The \textsc{resultative} subtype is narrow in scope, focusing specifically on causal/transformational relations between entities, which are frequently seen in contexts involving product-producing processes, such as cooking/baking (e.g., \underline{some flour} → \textbf{the bread}, \citealt{fang-etal-2022-take}), but not common in most contexts. The fact that instances of this subtype are uniformly predicted as the broader \textsc{associative} subtype indicates that \textsc{resultative} likely needs to be further detailed with more examples in the subtype classification prompt.

\section{Conclusion}


In this paper, we present LLMBridge, the first LLM based pipeline for bridging resolution. We evaluate our bridging resolution system in the challenging \textit{End-to-end} evaluation setting, along with the \textit{Basic} evaluation setting. We report scores on all three of the referential bridging evaluation datasets for English: ISNotes, BASHI, and GUMBridge. We additionally provide the first set of scores in the \textit{End-to-End} evaluation setting for the GUMBridge corpus on bridging resolution and bridging subtype classification. We achieve new SoTA results on all three datasets in both the \textit{End-to-end} and \textit{Basic} evaluation settings. We also provide error analysis of LLMBridge's performance on the evaluation data, finding that the system struggles to identify long distance instances of bridging as well as more abstract bridging subtypes, like \textsc{entity-associative}. 
Overall, LLMBridge produces strong results on all three of the referential bridging corpora for English, achieving new SoTA results across the board, including the highest ever scores reported on any English dataset for anaphor recognition and anaphor resolution in the \textit{End-to-end Setting}.

\section*{Limitations}

This paper focuses on creating a bridging resolution pipeline for English and does not explore the cross-linguistic differences in how bridging manifests. Future work is required to determine to what extent this pipeline could directly apply or be adapted for bridging in a multilingual setting. 
Additionally, the design of this pipeline focuses specifically on the anaphoric, information status based definition of bridging known as referential bridging. Further adaptation of the pipeline would be necessary to increase performance on lexical bridging. 
Also, results from previous work are reported rather than reproduced due to incomplete or unreproducible software repositories. This means we were not able to obtain p-values for differences between our results and previous numbers. 

We also acknowledge that the cost of running the LLMBridge pipeline with a SoTA commercial LLM backend (such as \texttt{Gemini-3.1-pro-preview}) is a significant expense, though what is an expensive SoTA model today is likely to rapidly reduce in cost, based on past experience. The cost of the evaluation conducted in this paper was approximately 1000 USD (500 USD for a single run of all models on all datasets), nearly all of which was for the results obtained from running LLMBridge with \texttt{Gemini-3.1-pro-preview}, which is a very new and expensive model. These costs limit both our ability to conduct multiple runs for evaluation and the ability for future work to reproduce our results, at least until a corresponding model becomes more affordable. While further exploration is required to find a balancing point between backend model cost and pipeline performance, results from this study indicate that LoRA fine-tuning a mid-sized LLM may be a promising direction, and the advantages of the largest available model at the time of writing are reflected in the scores we report in our results.


\bibliography{custom}

\appendix

\section{GUMBridge Bridging Subtypes}
\label{sec:appendix_subtype}

\begin{figure}[h]
  \centering
  \includegraphics[width=1\columnwidth]{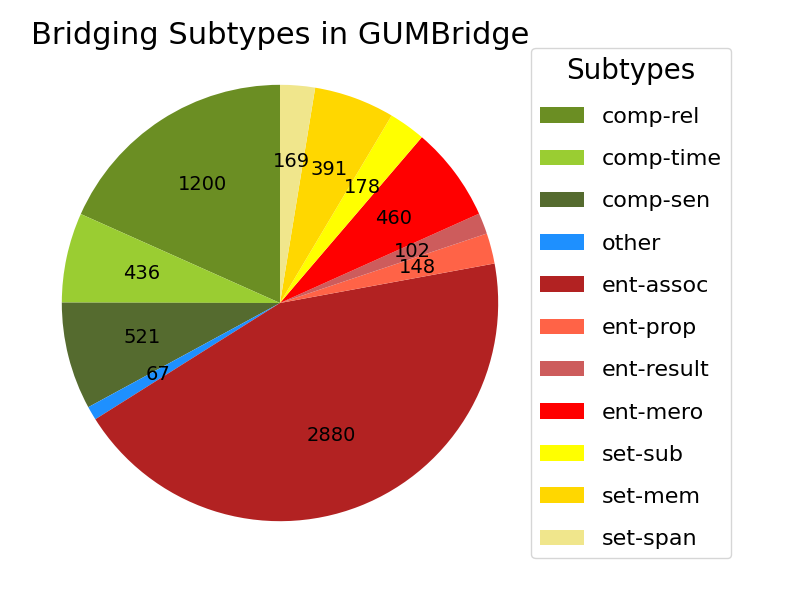}
  \caption{Bridging subtype proportions in the GUMBridge corpus. Raw counts for each subtype annotation are shown in the Figure.}
  \label{fig:subtype_proportions}
\end{figure}

This appendix briefly details the bridging subtype varieties annotated in the GUMBridge corpus. Figure \ref{fig:subtype_proportions} shows the proportions and raw counts of the subtype annotations in the GUMBridge corpus.

\paragraph{\textsc{comparison-relative}} The anaphor is preceded by a comparative marker which implies a comparison to the antecedent (e.g., \underline{several women} → \textbf{other women}). 

\paragraph{\textsc{comparison-sense}} The type of the anaphor is omitted but inferable via comparison to the antecedent (e.g., \underline{a Chinese restaurant} → \textbf{the Italian one}).

\paragraph{\textsc{comparison-time}} The anaphor refers to a specific time/time frame which is understandable with reference to the time/time frame expressed by the antecedent (e.g., \underline{Wednesday} → \textbf{yesterday}). 

\paragraph{\textsc{entity-meronomy}} The anaphor has a part-whole relation with the antecedent, including physical subparts, substance-portion, and regions/subsections (e.g., \underline{a house} → \textbf{the door}).

\paragraph{\textsc{entity-property}} The anaphor is a physical or intangible property of the antecedent, such as smell, length, or style (e.g., \underline{a bouquet of roses} → \textbf{the scent}).

\paragraph{\textsc{entity-resultative}} The anaphor is logically inferable from the antecedent. This is often the result of a transformative/product producing process, like cooking/baking (e.g., \underline{flour} → \textbf{the bread}).


\paragraph{\textsc{entity-associative}} The anaphor is an attribute or closely associated entity of the antecedent (e.g., \underline{a library} → \textbf{the books}). 

\paragraph{\textsc{set-member}} The anaphor is an element of the antecedent set. This includes group-member and class-instance relations (e.g., \underline{several books} → \textbf{the mystery novel}).

\paragraph{\textsc{set-subset}} The anaphor is a subset of the antecedent set (e.g., \underline{a group of students} → \textbf{the boys}).

\paragraph{\textsc{set-span-interval}} The anaphor is a sub-span of the spatial or temporal antecedent interval (e.g, \underline{Sunday} → \textbf{the morning}). 

\paragraph{\textsc{other}} The  \textsc{other} category is for instances which fit the information status based definition of a bridging pair but do not fall into any of the bridging subtype categories outlined above. 

\section{LLMBridge Pipeline Prompts}
\label{sec:appendix_prompts}


The following are examples of the prompt templates we use to query the LLM backend of LLMBridge for each bridging resolution subtask. 
To develop our LLM prompts for each bridging subtask, we utilized a broad definition of referential bridging that prioritizes high recall across datasets.
While we explored using more specific prompts specialized for each dataset definition of bridging, the resulting lower recall did not outweigh gains in precision.\footnote{During prompt development, we evaluated against a subset documents from the GUMBridge dev set (GUM\_academic\_librarians, GUM\_conversation\_grounded, GUM\_court\_loan, GUM\_bio\_emperor, GUM\_voyage\_athens), as well as dev documents we set aside from ISNotes (wsj\_1450, wsj\_1327, wsj\_1232, wsj\_1163, wsj\_1148) and BASHI (wsj\_1846, wsj\_2112, wsj\_1505, wsj\_0242, wsj\_0790).} As such, our finalized prompts use the more open definition of referential bridging combined with subtype explanations from the GUMBridge categorization schema and dataset specific few-shot examples for the individual subtasks.  

\paragraph{Anaphor Recognition}

\begin{tcolorbox}[promptbox]
\scriptsize

You are a linguistic assistant for identifying bridging anaphora. A bridging anaphor is a newly introduced entity whose referent in the context is inferable based on a relationship to a previous entity or verbal predicate in the discourse (the antecedent). For instance,  "a house … the door" (antecedent=a house, anaphor=the door) where we understand that the door is specifically the door of the house, and we cannot interpret the referent of the door without referring to the house. \\

There are a few types of such anaphora: \\
 
·  Comparison \\

comparison-relative: The anaphor is preceded by a comparative marker (other, another, same, more, ordinal modifiers, comparative adjectives, superlatives, etc.) which implies a comparison to the antecedent. For example: "The children ... another child" (=another with comparison to the aforementioned children); similar cases may be similar children, older children (compared to the aforementioned children), etc. \\

comparison-sense: the semantic type of a phrase requires a previous mention to identify it, for example "the Italian restaurant ... a Chinese one" (we can't know "a Chinese one" is a restaurant without referring back to the Italian restaurant), or "another one", "the others" etc. \\

comparison-time: the anaphor refers to a specific time/timeframe which is understandable with reference to the antecedent, for example: "Tuesday, February 2nd ... the following week" \\ 

·  Entity \\

entity-meronomy: the anaphor is a subunit of the antecedent (part-whole), including physical subunits, portion-substance relations, and regions/subsections. For example: "the house ... the door" (=of the house). \\

entity-associative: the anaphor is an attribute or closely associated entity of the antecedent, including both prototypical and inducible associations: "a wedding ... the bride" (=the bride at that wedding), implicit arguments of a predicate or a verbal nominalization: "a play... the performance" (=of the play), relational nouns: "a murder ... the victim" \\

entity-property: the anaphor is a physical or intangible property of the antecedent (e.g., smell, length, size, style, etc.): "the tea ... the sweet aroma" \\

entity-resultative: the anaphor is logically inferable from the antecedent (e.g., result, transformation/transmutation, cause): "the dough ... the bread" (=the dough becomes bread after baking) \\

·  Set \\

set-member: the anaphor is an element of the antecedent set, including groups-member relations and classes-instances: "the cars ... the Mazda", additionally indefinite members to definite sets: "a candle on each cupcake ... the candles" \\

set-subset: the anaphor is a subset of the antecedent set: "the cars ... the Mazdas" (not all Mazdas, just the subset among the aforementioned cars) \\

set-span-interval: the anaphor is a sub-span of a spatial or temporal interval defined by the antecedent: "last week ... Wednesday" (=Wednesday of last week), "Sunday ... the morning" (=the morning portion of that Sunday) \\

TASK DEFINITION \\
Given a larger passage and a specified subspan labeled “Text: …”, return a list of all bridging anaphora (newly introduced entities whose referent is inferable based on a relationship to a previous entity or verbal predicate). Output must be the exact surface strings from the subspan. If none, return []. \\

OUTPUT CONSTRAINTS (STRICT) \\
- Return a JSON-style list of exact strings from the text. \\
- If no predicted bridging anaphora, return: [] \\
- Do not add explanations or any extra text. \\

CRITICAL REMINDERS \\
- The bridging anaphor must be a newly introduced entity (it cannot corefer with anything the occurs before it in the text) \\
- The interpretation of the bridging anaphor MUST depend on a PREVIOUS entity or verbal predicate for interpretation \\
- Return bridging anaphora as contiguous mention span exactly as written in the text (including unusual spacing, hyphenation, parentheses tokens like -LRB-/-RRB-, and any trailing comma that is part of the noun phrase). \\
- Entities should be considered with their full phrases – if a noun is expanded by modifier clauses etc., include the entire noun phrase (maximal projection),  e.g., in "the man who I saw", not just "the man" \\

TASK EXAMPLES \\
\{dataset\_specific\_examples\} \\

TASK \\
Consider the following text: \\
\{context\_text\} \\
Please return a list of all bridging anaphora in the following subspan of the text. \\
Text: \\
\{text\} \\
Answer(s): \\
\end{tcolorbox}

\paragraph{Anaphor Recognition with Candidates}

\begin{tcolorbox}[promptbox]
\scriptsize

You are a linguistic assistant for identifying bridging anaphora. A bridging anaphor is a newly introduced entity whose referent in the context is inferable based on a relationship to a previous entity or verbal predicate in the discourse (the antecedent). For instance,  "a house … the door" (antecedent=a house, anaphor=the door) where we understand that the door is specifically the door of the house, and we cannot interpret the referent of the door without referring to the house. \\

There are a few types of such anaphora: \\
 
·  Comparison \\

comparison-relative: The anaphor is preceded by a comparative marker (other, another, same, more, ordinal modifiers, comparative adjectives, superlatives, etc.) which implies a comparison to the antecedent. For example: "The children ... another child" (=another with comparison to the aforementioned children); similar cases may be similar children, older children (compared to the aforementioned children), etc. \\

comparison-sense: the semantic type of a phrase requires a previous mention to identify it, for example "the Italian restaurant ... a Chinese one" (we can't know "a Chinese one" is a restaurant without referring back to the Italian restaurant), or "another one", "the others" etc. \\

comparison-time: the anaphor refers to a specific time/timeframe which is understandable with reference to the antecedent, for example: "Tuesday, February 2nd ... the following week" \\ 

·  Entity \\

entity-meronomy: the anaphor is a subunit of the antecedent (part-whole), including physical subunits, portion-substance relations, and regions/subsections. For example: "the house ... the door" (=of the house). \\

entity-associative: the anaphor is an attribute or closely associated entity of the antecedent, including both prototypical and inducible associations: "a wedding ... the bride" (=the bride at that wedding), implicit arguments of a predicate or a verbal nominalization: "a play... the performance" (=of the play), relational nouns: "a murder ... the victim" \\

entity-property: the anaphor is a physical or intangible property of the antecedent (e.g., smell, length, size, style, etc.): "the tea ... the sweet aroma" \\

entity-resultative: the anaphor is logically inferable from the antecedent (e.g., result, transformation/transmutation, cause): "the dough ... the bread" (=the dough becomes bread after baking) \\

·  Set \\

set-member: the anaphor is an element of the antecedent set, including groups-member relations and classes-instances: "the cars ... the Mazda", additionally indefinite members to definite sets: "a candle on each cupcake ... the candles" \\

set-subset: the anaphor is a subset of the antecedent set: "the cars ... the Mazdas" (not all Mazdas, just the subset among the aforementioned cars) \\

set-span-interval: the anaphor is a sub-span of a spatial or temporal interval defined by the antecedent: "last week ... Wednesday" (=Wednesday of last week), "Sunday ... the morning" (=the morning portion of that Sunday) \\

TASK DEFINITION \\
Given a larger passage, a candidate list of mentions, and a specified subspan labeled “Text: …”, return a list of all bridging anaphora (newly introduced entities whose referent is inferable based on a relationship to a previous entity or verbal predicate). Output must be the exact surface strings from the candidate list. If none, return []. \\

OUTPUT CONSTRAINTS (STRICT) \\
- Return a JSON-style list of exact strings from the candidate list. \\
- If no candidates are bridging, return: [] \\
- Do not add explanations or any extra text. \\

CRITICAL REMINDERS \\
- The bridging anaphor must be a newly introduced entity (it cannot corefer with anything the occurs before it in the text) \\
- The interpretation of the bridging anaphor MUST depend on a PREVIOUS entity or verbal predicate for interpretation \\
- Return bridging anaphora as contiguous mention span exactly as written in the candidate list (including unusual spacing, hyphenation, parentheses tokens like -LRB-/-RRB-, and any trailing comma that is part of the noun phrase). \\
- Entities should be considered with their full phrases – if a noun is expanded by modifier clauses etc., include the entire noun phrase (maximal projection),  e.g., in "the man who I saw", not just "the man" \\

TASK EXAMPLES \\
\{dataset\_specific\_examples\} \\

TASK \\
Consider the following text: \\
\{context\_text\} \\
Please return a list of all bridging anaphora in the following subspan of the text. \\
Text: \\
\{text\} \\
Candidate Anaphora: \\
\{candidate\_anaphor\_list\} \\
Answer(s): \\
\end{tcolorbox}

\paragraph{Anaphor Resolution}

\begin{tcolorbox}[promptbox]
\scriptsize
You are a linguistic assistant tasked with finding the antecedent of a bridging anaphor. A bridging anaphor is a newly introduced entity whose referent in the context is inferable based on a relationship to a previous entity or verbal predicate in the discourse (the antecedent). For instance,  "a house … the door" (antecedent=a house, anaphor=the door) where we understand that the door is specifically the door of the house, and we cannot interpret the referent of the door without referring to the house. The antecedent is the entity of verabl predicate that allows the reader to understand the referent of the bridging anaphor. \\

There are a few types of such anaphora with their antecedents: \\
 
·  Comparison \\

comparison-relative: The anaphor is preceded by a comparative marker (other, another, same, more, ordinal modifiers, comparative adjectives, superlatives, etc.) which implies a comparison to the antecedent. For example: "The children ... another child" (=another with comparison to the aforementioned children, antecedent=The children); similar cases may be similar children, older children (compared to the aforementioned children), etc. \\

comparison-sense: the semantic type of a phrase requires a previous mention to identify it, for example "the Italian restaurant ... a Chinese one" (we can't know "a Chinese one" is a restaurant without referring back to the Italian restaurant, antecedent=the Italian restaurant), or "another one", "the others" etc. \\

comparison-time: the anaphor refers to a specific time/timeframe which is understandable with reference to the antecedent, for example: "Tuesday, February 2nd ... the following week" (antecedent=Tuesday, February 2nd) \\

·  Entity \\

entity-meronomy: the anaphor is a subunit of the antecedent (part-whole), including physical subunits, portion-substance relations, and regions/subsections. For example: "the house ... the door" (=of the house, antecedent=the house). \\

entity-associative: the anaphor is an attribute or closely associated entity of the antecedent, including both prototypical and inducible associations: "a wedding ... the bride" (=the bride at that wedding, antecedent=a wedding), implicit arguments of a predicate or a verbal nominalization: "a play... the performance" (=of the play, antecedent=the play), relational nouns: "a murder ... the victim" (antecedent=a murder) \\

entity-property: the anaphor is a physical or intangible property of the antecedent (e.g., smell, length, size, style, etc.): "the tea ... the sweet aroma" (antecedent=the tea) \\

entity-resultative: the anaphor is logically inferable from the antecedent (e.g., result, transformation/transmutation, cause): "the dough ... the bread" (=the dough becomes bread after baking, antecedent=the dough) \\

·  Set \\

set-member: the anaphor is an element of the antecedent set, including groups-member relations and classes-instances: "the cars ... the Mazda" (antecedent=the cars), additionally indefinite members to definite sets: "a candle on each cupcake ... the candles" (antecedent=a candle on each cupcake) \\

set-subset: the anaphor is a subset of the antecedent set: "the cars ... the Mazdas" (not all Mazdas, just the subset among the aforementioned cars, antecedent=the cars) \\

set-span-interval: the anaphor is a sub-span of a spatial or temporal interval defined by the antecedent: "last week ... Wednesday" (=Wednesday of last week, antecedent=last week), "Sunday ... the morning" (=the morning portion of that Sunday, antecedent=Sunday) \\

TASK DEFINITION \\
You will be given a text with a possible anaphor marked in double curly brackets: \{\{ \}\}. Output exactly the string of the antecedent (if there is one), or 'no antecedent' (if there is no antecedent in the given text). If there are multiple mentions of the antecedent, select the one closest to (but still before) the anaphor, even if it is a pronoun. For example in: "the house … it … \{\{the door\}\}", if "it" refers to the house, the correct solution is: it \\

OUTPUT CONSTRAINTS (STRICT) \\
- Return exactly one string: the antecedent mention copied verbatim from the text. \\
- The string must be a single, contiguous span that: \\
  - Appears before the marked anaphor \\
  - Is not identical to the anaphor and is not coreferential with it \\
- Do not add explanations, quotes, brackets, or multiple spans. \\
- If no associative antecedent exists in the text (rare), return exactly: no antecedent \\

CRITICAL REMINDERS \\
- The antecedent must precede the anaphor. \\
- Return a single contiguous mention span exactly as written (including unusual spacing, hyphenation, parentheses tokens like -LRB-/-RRB-, and any trailing comma that is part of the noun phrase). \\
- Return full phrases – if a noun is expanded by modifier clauses etc., include the entire noun phrase (maximal projection) \\
- Do not return a paraphrase or combine multiple spans. \\
- If no specific prior mention is required to interpret the anaphor, output: 'no antecedent' \\

TASK EXAMPLES \\
\{dataset\_specific\_examples\} \\

TASK \\
Please return a single string for the associative antecedent of the bridging anaphor surrounded by double curly brackets: \{\{ \}\}. \\
Text: \\
\{text\} \\
Answer: \\
\end{tcolorbox}

\paragraph{Subtype Classification}

\begin{tcolorbox}[promptbox]
\scriptsize
TASK DEFINITION \\
Classify the subtype(s) of bridging relation between one marked anaphor and its corresponding marked antecedent. \\

INPUT FORMAT \\
- You are given a text containing one bridging anaphor–antecedent pair \\
	- the anaphor, marked with double curly brackets: \{\{ \}\} \\
	- the antecedent, marked with asterisks: * * \\
- The relation to classify is specifically between the starred antecedent mention and the double-braced anaphor, using the surrounding context as needed. \\

WHAT COUNTS AS A BRIDGING ANAPHOR \\
- A newly introduced noun phrase (NP) whose interpretation depends on a previously mentioned but non-identical entity or verbal predicate (the antecedent). \\
- The anaphor does not corefer with the antecedent. \\

DECISION PROCEDURE (IN ORDER) \\
1) Identify the semantic relation(s) needed to interpret the anaphor given the antecedent. \\
2) Assign all applicable subtypes that are directly licensed by the text and conventional world knowledge (e.g., typical roles/participants, frames). \\
   - Multiple subtypes may apply; include all that fit. \\
3) If none of the defined subtypes apply, output "other". \\

SUBTYPE DEFINITIONS (WITH DISAMBIGUATION RULES) \\

Comparison based \\
- comparison-relative: The anaphor is introduced with a comparative/superlative/ordinal or related marker (e.g., another, other, same, more, most, less, fewer, similar, different, next, last, first, second, better, best, worse, worst). \\
  - Includes temporal NPs like “the last time,” “the next day” when they are picked out relative to a previously mentioned event/situation. \\
- comparison-sense: The anaphor’s type/kind is recovered from the antecedent mention (e.g., “one/ones/others” whose category is supplied by the antecedent; or an NP whose category is understood from the prior mention). \\
  - Often co-occurs with comparison-relative when both a comparative marker and type-recovery are present (e.g., “the others” relative to a prior set; “the last time” where “time of [that event]” is understood from context). \\
- comparison-time: Use only for temporal anchoring across intervals (e.g., a shift from a previously mentioned specific time span to a following/preceding larger or adjacent interval). Example: “Tuesday” → “the following week”. \\
  - Do NOT use for ordinal/superlative temporal NPs like “the last time” anchored to a prior event; prefer comparison-relative (and add comparison-sense if the event/time type comes from the antecedent). \\

Entity based \\
- entity-meronomy: Part–whole relations (physical subparts, regions/subsections, portion–substance). \\
  - Examples: “the house” → “the door”; “the cake” → “a slice”. \\
- entity-associative: Prototypical associations, roles, attributes, frames, or implicit arguments tied to the antecedent entity or event. \\
  - Examples: “a wedding” → “the bride”; “a play” → “the performance”; a person → their typical activities like “work” or “sleep”. \\
- entity-property: A property or attribute of the antecedent (physical or intangible), e.g., smell, size, style, mood. \\
  - Example: “the tea” → “the sweet aroma”. \\
- entity-resultative: Result/cause or transformation relations (inputs → outputs or vice versa). \\
  - Example: “the flour” → “the bread”. \\

Set based \\
- set-member: Group–member or class–instance relations. \\
  - Examples: “the cars” → “the Mazda”; “mammals” → “a whale”. \\
- set-subset: A subset picked from a previously mentioned set. \\
  - Example: “the cars” → “the Mazdas”. \\
- set-span-interval: A sub-span within a previously defined spatial or temporal interval. \\
  - Example: “last week” → “Wednesday”. \\
  - Include this with comparison-time when the anaphor is a clear sub-part of an explicit interval named by the antecedent. \\

Other \\
- Use “other” only when none of the above categories fit but interpretation still depends on the antecedent. \\

KEY DISAMBIGUATION GUIDELINES \\
- Comparison vs. time: \\
  - Ordinal/comparative temporal expressions like “the last time,” “the next day” → comparison-relative and comparison-time. If their type (“time of that event”) comes from the prior mention, also add comparison-sense. \\
  - Cross-interval anchoring like “Tuesday” → “the following week” → comparison-relative and comparison-time. \\
- Type-recovery: \\
  - “one/ones/others” whose category is supplied by the antecedent → comparison-sense (often also comparison-relative if a comparative marker is present). \\
- Prototypical associations: \\
  - Daily routines/activities linked to a person or role (e.g., “A contemporary American” → “work”; “you” → “sleep”) → entity-associative. \\
- Set vs. meronymy: \\
  - Member-of-a-set/class → set-member. Physical part-of → entity-meronomy. \\
- Use all applicable labels when multiple relations are licensed. \\

OUTPUT CONSTRAINTS (STRICT) \\
- Output only the label(s). \\
- If multiple labels apply, join them with semicolons and NO SPACES. Example: comparison-sense;comparison-relative \\
- Do not add explanations or any other text. \\
- Do not invent labels. \\
- Allowed labels (exact spellings): \\
  comparison-relative \\
  comparison-sense \\
  comparison-time \\
  entity-associative \\
  entity-meronomy \\
  entity-property \\
  entity-resultative \\
  set-member \\
  set-subset \\
  set-span-interval \\
  other \\

WORKED EXAMPLE PATTERNS (FOR CONSISTENCY) \\
- Antecedent: prior event/situation; Anaphor: “the last time” → comparison-sense;comparison-relative \\
- Antecedent: “A contemporary American”; Anaphor: “work” → entity-associative \\
- Antecedent: “you”; Anaphor: “sleep” → entity-associative \\

TASK EXAMPLES \\
\{dataset\_specific\_examples\} \\

TASK \\
In the following text, a bridging anaphor is marked with double curly brackets \{\{ \}\}, and the corresponding antecedent is surrounded by asterisks: * * . \\
Classify the subtype(s) of bridging relation that hold between the two entities. \\
Text: \\
\{text\} \\
Answer: \\
\end{tcolorbox}

\section{Impact of Pre/Post Processing}
\label{sec:processing_comparison}

Table \ref{tab:no_processing_results} shows a two run average of the bridging resolution results for LLMBridge (\texttt{Qwen2.5-7B-Instruct-FT} backend) without pre/post processing. Comparing these results to those in Table \ref{tab:end2end_results} and Table \ref{tab:reg_results}, which do leverage the heuristic pre/post processing of the pipeline, we see that performance on anaphor recognition and resolution is degraded, indicating that the pre/post processing heuristics are valuable (particularly for precision). In the \textit{End-to-end Setting} we see an avg. $\Delta$ F1 across datasets of $-7.7$ for anaphor recognition and $-5.0$ for anaphor resolution. In the \textit{Basic Setting} there is an avg. $\Delta$ accuracy of $-3.6$ for anaphor resolution. 

\begin{table}[h]
\centering
\resizebox{\columnwidth}{!}{%
\begin{tabular}{cccc}
\toprule
\multicolumn{4}{c}{\textbf{End-to-end Bridging Resolution Setting}} \\ \midrule
\multirow{2}{*}{} & \multicolumn{3}{c}{Recognition} \\
 & P & R & F1 \\ \midrule
\textbf{GUMBridge} & 23.0 $\pm$ 0.6  & 36.0 $\pm$ 0.1 & 28.0 $\pm$ 0.5 \vspace{0.1cm} \\
\textbf{ISNotes} & 18.3 $\pm$ 0.3 & 27.5 $\pm$ 0.4 & 21.9 $\pm$ 0.1 \vspace{0.1cm} \\
\textbf{BASHI} & 10.7 $\pm$ 0.2 & 39.1 $\pm$ 0.2 & 16.8 $\pm$ 0.3 \\ \midrule
\multirow{2}{*}{} & \multicolumn{3}{c}{Resolution} \\
 & P & R & F1 \\ \midrule
\textbf{GUMBridge} & 6.7 $\pm$ 0.2 & 10.5 $\pm$ 0.0 & 8.2 $\pm$ 0.1 \vspace{0.1cm} \\
\textbf{ISNotes} & 4.4 $\pm$ 0.0 & 6.7 $\pm$ 0.2 & 5.3 $\pm$ 0.1 \vspace{0.1cm} \\
\textbf{BASHI} & 1.5 $\pm$ 0.1 & 5.7 $\pm$ 0.2 & 2.4 $\pm$ 0.1 \\ \midrule
\multicolumn{4}{c}{\textbf{Basic Bridging Resolution Setting}} \\ \midrule
 & \multicolumn{3}{c}{Accuracy} \\ \midrule
\textbf{GUMBridge} & \multicolumn{3}{c}{28.1 $\pm$ 0.3} \vspace{0.1cm} \\
\textbf{ISNotes} & \multicolumn{3}{c}{23.5 $\pm$ 0.1} \vspace{0.1cm} \\
\textbf{BASHI} & \multicolumn{3}{c}{21.0 $\pm$ 0.8} \\ \bottomrule
\end{tabular}%
}
\caption{LLMBridge (\texttt{Qwen2.5-7B-Instruct-FT} backend) bridging resolution results without pre/post processing in \textit{End-to-end} and \textit{Basic} evaluation settings.}
\label{tab:no_processing_results}
\end{table}

\end{document}